\documentclass[conference, letterpaper]{IEEEtran}

\IEEEoverridecommandlockouts

\usepackage{pgfplots}
\usepackage{multirow}
\usepackage{tabularx}
\usepackage{setspace}
\usepackage{pgfplots}
\usepackage{filecontents}
\usepackage{listings}
\usepackage{cite}
\usepackage{amsmath,amssymb,amsfonts, bm}
\usepackage{algorithm}
\usepackage[noend]{algorithmic}
\usepackage{graphicx,wrapfig}
\usepackage{graphicx}
\usepackage{textcomp}
\usepackage{xcolor}
\usepackage{balance}
\usepackage{chronosys}
\usepackage[T1]{fontenc}
\usepackage{comment}
\usepackage{soul}

\usepackage{physics}
\newcommand{\btheta}{\boldsymbol{\theta}}

\newcommand{\calL}{\mathcal{L}}
\newcommand{\calN}{\mathcal{N}}

\DeclareMathOperator*{\argmax}{arg\,max}
\DeclareMathOperator*{\argmin}{arg\,min}

\IEEEoverridecommandlockouts\IEEEpubid{\makebox[\columnwidth]{978-1-6654-3902-2/21/\$31.00 2021 $\copyright$ IEEE \hfill}\hspace{\columnsep}\makebox[\columnwidth]{ }}

\begin{document}

\title{Zone-based Federated Learning \\ for Mobile Sensing Data}

\author{
\IEEEauthorblockN{
 	Xiaopeng Jiang\IEEEauthorrefmark{1}
	Thinh On\IEEEauthorrefmark{1}
 	NhatHai Phan\IEEEauthorrefmark{1}
  Hessamaldin Mohammadi\IEEEauthorrefmark{1}
	Vijaya Datta Mayyuri\IEEEauthorrefmark{2}
	An Chen\IEEEauthorrefmark{2} }
	Ruoming Jin\IEEEauthorrefmark{3}
	Cristian Borcea\IEEEauthorrefmark{1}
\\
\IEEEauthorblockN{
	New Jersey Institute of Technology\IEEEauthorrefmark{1}
	Qualcomm Incorporated\IEEEauthorrefmark{2}
	Kent State University\IEEEauthorrefmark{3}
}

\IEEEauthorblockA{		
		Email:\{xj8,to58,phan,hm385,borcea\}@njit.edu, \{vmayyuri,anc\}@qualcomm.com, rjin1@kent.edu
		}
}

\maketitle
\begingroup\renewcommand\thefootnote{\textsection}
\endgroup

\begin{abstract}
Mobile apps, such as mHealth and wellness applications, can benefit from deep learning (DL) models trained with mobile sensing data collected by smart phones or wearable devices. However, currently there is no mobile sensing DL system that simultaneously achieves good model accuracy while adapting to user mobility behavior, scales well as the number of users increases, and protects user data privacy. We propose Zone-based Federated Learning (ZoneFL) to address these requirements. ZoneFL divides the physical space into geographical zones mapped to a mobile-edge-cloud system architecture for good model accuracy and scalability. Each zone has a federated training model, called a zone model, which adapts well to data and behaviors of users in that zone. Benefiting from the FL design, the user data privacy is protected during the ZoneFL training. We propose two novel zone-based federated training algorithms to optimize zone models to user mobility behavior: Zone Merge and Split (ZMS) and Zone Gradient Diffusion (ZGD). ZMS optimizes zone models by adapting the zone geographical partitions through merging of neighboring zones or splitting of large zones into smaller ones. Different from ZMS, ZGD maintains fixed zones and optimizes a zone model by incorporating the gradients derived from neighboring zones' data. ZGD uses a self-attention mechanism to dynamically control the impact of one zone on its neighbors. Extensive analysis and experimental results demonstrate that ZoneFL significantly outperforms traditional FL in two models for heart rate prediction and human activity recognition. In addition, we developed a ZoneFL system using Android phones and AWS cloud. The system was used in a heart rate prediction field study with 63 users for 4 months, and we demonstrated the feasibility of ZoneFL in real-life. 

\end{abstract}


\section{Introduction}

Sensing data collected on mobile devices can be employed in many novel Deep Learning (DL) applications. A practical application domain is mobile health and wellness, where data is collected by smart phones with potential help from wearable sensors. An effective DL model requires data from many users, and it needs to protect the privacy of sensitive mobile sensing data. Furthermore, it needs to adapt to user behavior, which is location-dependent. For example, people's lifestyles depend on their living areas. Living in dense areas of the city with fewer recreational facilities prevents people from doing enough exercise. Similarly, health problems may be related to the level of pollution in different parts of the city.

The aim of this paper is to build an effective DL system for mobile sensing data that works efficiently on smart phones and satisfies the following requirements: (i) \textit{Privacy-preserving}: learn from data provided by many users, while protecting user data privacy; (ii) \textit{Mobility-awareness}: achieve good model accuracy by adapting to user mobility behavior, and (iii) \textit{Scalability}: scale well as the number of users increases. We propose \textbf{Zone-based Federated Learning (ZoneFL)}, a novel federated learning (FL) architecture that builds and manages different models for different geographical zones, to satisfy these requirements. By design, ZoneFL satisfies the privacy-preserving requirement because Federated Learning (FL)~\cite{mcmahan2017fedAVG} learns from data collected by many users, while protecting the user data privacy during training. In FL, the models are trained on mobile devices with their local data, and the server aggregates the models received from mobile devices. The users' privacy-sensitive data never leave the mobile devices. 

We give vehicular traffic prediction and heart health notification as two concrete motivating examples. For traffic prediction, the traffic patterns in shopping districts and business districts are different because of different zone-dependent user behavior.  A heart health notification app sends alerts about the level of cardiovascular risk associated with users’ current activity based on the altitude and climate of a geographical zone. Using ZoneFL will outperform a global model in such applications, and we enjoy privacy prescerving and scalability of ZoneFL as well.

The main novelty of ZoneFL is its zone-based approach to satisfy requirements for mobility-awareness and scalability. To adapt DL models to user mobility for higher accuracy and to achieve good scalability, ZoneFL divides the physical space into geographically non-overlapping zones mapped to a mobile-edge-cloud architecture. Each zone trains its own zone model, which adapts to the data and behaviors of the users who  spend time in that zone. As users move from one zone to another, collect data, and participate the training of different zones. For inference, their mobile devices switch from one zone model to another. Thus, zone models achieve higher accuracy than globally trained FL models, satisfying the mobility-awareness requirement. In ZoneFL, edge nodes manage the FL training within their zones and host the latest models for their zones. Mobile devices can download these models when they enter a new zone. The cloud collaborates with the edge nodes to dynamically maintain the zone partitions for the entire space, but it is not involved in training. Compared to traditional FL mobile-cloud architecture, the mobile-edge-cloud architecture of ZoneFL is more scalable because model aggregation is done distributedly at the edge (satisfying the scalability requirement), has lower latency for mobile users who interact with the edge instead of the cloud, and results in less bandwidth consumption in the network core~\cite{abbas2017Mobile-edge-survey2,murshed2019machineEdgeSurvey}. 

A major challenge in ZoneFL is how to ensure the zone models adapt to user mobility behavior changes over time. To solve this challenge, we propose two novel zone-based federated training algorithms: Zone Merge and Split (ZMS) and Zone Gradient Diffusion (ZGD). ZMS optimizes zone models by adapting the zone geographical partitions through merging of neighboring zones or splitting of large zones back to previously merged smaller zones. The algorithm ensures that merging and splitting results in better model accuracy in each new zone. ZMS can be used when the initial zone partitions are suboptimal, and the zone partitions will be gradually improved as ZMS proceeds. Different from ZMS, ZGD maintains fixed zones and optimizes a zone model by leveraging concepts from graph neural networks to incorporate the gradients derived from neighboring zones' data. ZGD uses a self-attention mechanism to dynamically control the impact of one zone on its neighbors. ZGD can be used to further optimize zone models when the zone partitions are relatively stable according to ZMS.

ZoneFL was evaluated in terms of model accuracy and system performance using two models and two real-world datasets: Human Activity Prediction (HAR) with mobile sensing data collected in the wild, and Heart Rate Prediction (HRP) with the FitRec dataset~\cite{ni2019modeling}~\footnote{Datasets were downloaded and evaluated by the NJIT team.}. The results demonstrate that models using ZoneFL without optimization performed by ZMS and ZGD significantly outperform their counterpart models using traditional FL for zones that have enough training data. ZoneFL with ZGD and ZMS further imporve the model performance, with ZMS improving the performance in the initial rounds and ZGD after that.  

We implemented a ZoneFL system using Android phones and AWS cloud. The system was tested with the HRP model in a field study in the wild with 63 users for 4 months. The results show that ZoneFL achieves low training and inference latency, as well as low memory and battery consumption on the phones.  ZoneFL scales better, because a zone edge server only handles only 34.98\% to 37.26\% of the communication and computation load handled by a global FL server. We also observed multiple zone merges and splits in the field study, when the model utility improved significantly. Compared with global FL, ZoneFL has a slightly higher training time on the mobile phones when the users participate in training for several zones. This overhead is an acceptable cost for the benefits provided by ZoneFL. Overall, the system results demonstrate the feasibility of ZoneFL in a real-life deployment.

The rest of this paper is organized as follows. Section~\ref{sec:related} reviews the related work. Section~\ref{sec:training} presents the ZoneFL training and the algorithms to dynamically adapt to user mobility.  Section~\ref{sec:arch} describes the design and implementation of the ZoneFL system. Section~\ref{sec:exp} shows the experimental results and analysis. The paper concludes and discusses future work in Section~\ref{sec:con}.

\section{Related Work}
\label{sec:related}


\subsection{Federated Learning}

Federated learning (FL) is a multi-round communication protocol between a coordination server and a set of $N$ clients to jointly train a learning model $f_{\theta}$, where $\theta$ is a vector of model parameters (also called weights). The training proceeds in rounds. At each round $t$ the server sends the latest model weights $\theta_t$ to a randomly sampled subset of clients $S_t$. Upon receiving $\theta_t$, each client $u \in S_t$ uses $\theta_t$ to train its local model and generates model weights $\theta^u_t$. Client $u$
computes its local gradient $\nabla \theta^u_t = \theta^u_t - \theta_t$, and sends it back to the server. After receiving the local gradients from all the clients in $S_t$, the server updates the model weights by aggregating all the received local gradients using an aggregation function $\mathcal{G}: \mathbb{R}^{|S_t|\times n} \rightarrow \mathbb{R}^n$, where $n$ is the size of $\nabla \theta^u_t$. The aggregated gradient will be added to $\theta_t$: $\theta_{t+1} = \theta_t + \lambda \mathcal{G}(\{\nabla\theta^i_t\}_{i\in S_t})$, where $\lambda$ is the server’s learning rate. A typical and widely applied aggregation function $\mathcal{G}$ is the weighted averaging, called Federated Averaging (FedAvg) \cite{kairouz2019advances}.


By joining the FL protocol, clients minimize the average of their loss functions as follows:
$\theta^* = \arg\min_\theta \frac{1}{N} \sum_{u = 1}^N \mathcal{L}_u(\theta)$, where $\mathcal{L}_u$ is the loss function of client $u$ on their local training dataset $D_u$. $\mathcal{L}_u$ is defined as $\mathcal{L}_u(\theta) = \frac{1}{|D_u|}\sum_{x \in D_u} \mathcal{L}\big(f_{\theta}(x), y \big)$, where $|D_u|$ denotes the number of data samples in $D_u$, and $\mathcal{L}$ is a loss function (e.g., cross-entropy) penalizing the mismatch between the predicted values $f_\theta(x)$ of an input $x$ and its associated ground-truth label $y$.

\textbf{Location Embedding in FL.} To adapt to user mobility behavior, a naive approach in FL could be to incorporate the user location in the model input~\cite{jiang2021federated, sprague2018asynchronous,li2020predicting}. However, compared with a model without location input, such an approach increases both the model size and the computation overhead, which leads to extra resource consumption on the mobiles. Different from these approaches, ZoneFL balances the trade-offs between model utility and system scalability by developing novel federated training algorithms seamlessly integrated into a scalable mobile-edge-cloud system architecture. 
Furthermore, potential attacks by an honest-but-curious server in an FL system that embeds user locations may be able to infer user mobility traces from the model weights. In ZoneFL, such a location privacy breach is more difficult because the fine-grained user location is not embedded in the models. 


\subsection{Clustering and Personalization in FL} 


As FL being adapted in pervasive computing~\cite{9439129, usmanova2021distillation}, user clustering has been proposed to improve the model accuracy of traditional FL. 
Clustering in FL~\cite{briggs2020federated,presotto2022fedclar} groups clients by the similarity of their local updates and trains the clusters independently. MLMG~\cite{9431011} uses a Multi-Local and Multi-Global model aggregation to train the non-IID user data with clustering methods. Clustered FL~\cite{sattler2020clustered} performs clustering with geometric properties of the FL loss surface. However, these works have the same scalability issue as traditional FL because they require a central server to cluster users. Khan et al.~\cite{khan2020self} propose an FL scheme with a clustering algorithm based on social awareness, which selects cluster heads to avoid a centralized server. In~\cite{he2019single-sidedFederated}, users share their model parameters with a group of trusted friends. One problem with these solutions is that utilizing social relationships to create clusters carries privacy risks.  

Although ZoneFL shares the idea of training models over groups of users with clustering approaches, there is no efficient clustering method to group users by their mobility behavior without violating users' location privacy. ZoneFL optimizes models to user mobility behavior and does not require centralized model updates or privacy-sensitive user information. The edge managers do not have access to users' locations; they just know that the user has been in a possibly large zone. 
Furthermore, ZoneFL
provides a solution that can be naturally deployed at the edge for better scalability, which is a further advantage compared to clustering approaches.

Personalized FL can also improve the FL model performance by mitigating the issue of non-independent and identically distributed (non-IID) data, which leads to lower performance in FL compared to centralized learning. Its key idea is to learn a personalized model per user~\cite{mansour2020three}. There are different methods for adapting global models for individual users~\cite{kulkarni2020survey}, including adding user context, transfer learning, using personalized layers, knowledge distillation, etc. Ditto~\cite{li2021ditto} leverages global-regularized multi-task learning to provide fairness and robustness through personalization in FL. In the adaptive personalized FL~\cite{deng2020adaptive}, each user trains a local model incorporating certain mixed weights in the global model. 
Ozkara et al.~\cite{ozkara2021quped} use quantization and distillation for personalized compression in FL. Although effective, these solutions demand extra computation on mobiles, which may negatively affect their resource consumption. 
ZoneFL is orthogonal to personalized FL, which can be leveraged in ZoneFL to produce personalized models for each user in each zone.


\subsection{Deep Learning at the Edge} 

DL has been employed in edge computing for a broad range of applications, such as video analytics, speech recognition, and autonomous vehicles~\cite{Mledge1,Mledge2,lin2018Mledge3,Mledge4,Mledge5}. Most of these works do not tackle the problem of privacy. Cui et al.~\cite{Stochastic} presented an online method to learn network changes and increase the network throughput. Bouazizi et al.~\cite{bouazizi2022low} proposed using low-resolution infrared array sensors to identify the presence and location of people indoors using edge DL. These works simply offload data collected by a variety of sensors to the edge, and can benefit from the privacy offered by ZoneFL. 

Several efforts have been carried out to leverage the computation power of both the cloud and the edge for DL. 
In~\cite{li2018learningIotinEdge}, the layers of a deep learning network are divided between edge servers and the cloud, and~\cite{DistributedDeepNeural} proposed a distributed architecture over an edge-cloud infrastructure. Part of the models and tasks in these works are executed in the cloud. These solutions still collect device data to the edge, and hence present privacy problems. Furthermore, unlike these solutions, ZoneFL achieves better scalability because the training is done at the edge, without any involvement from the cloud. 

\section{ZoneFL Training}
\label{sec:training}
This section presents zone partition, an overview of the ZoneFL training, and then describes our two federated training algorithms that allow ZoneFL to adapt to changes in user mobility.

\subsection{Zone Partition}

The physical space (e.g., a city) is partitioned into non-overlapping zones, based on administrative boundaries or other knowledge about their characteristics (e.g., shopping district, park, etc.). The zones are model-specific. For example, a heart rate prediction model has different zones compared with a vehicular traffic prediction model. In this way, ZoneFL can achieve better model performance by targeting training to zones in which the user behavior is more homogeneous for a given type of mobile sensing data. For example, the user mobility behavior in a park (e.g., exercising) is different from the behavior in a shopping districts (e.g., leisurely walking).

The granularity of zones can be defined based on the target application and the size of the user pool, i.e., each zone shall be small enough for behavior differences, while big enough to have sufficient users for better scalability-utility balance. The zone topology is a graph defined by neighboring relations of zones. By default the neighboring relation is adjacency (i.e., two zones are neighbors if their borders touch each other), but this could be modified, for example to define that two zones geographically closer than a given threshold are neighbors.

\begin{figure}[t!]
\centering
\includegraphics[width=1\linewidth,scale=1]{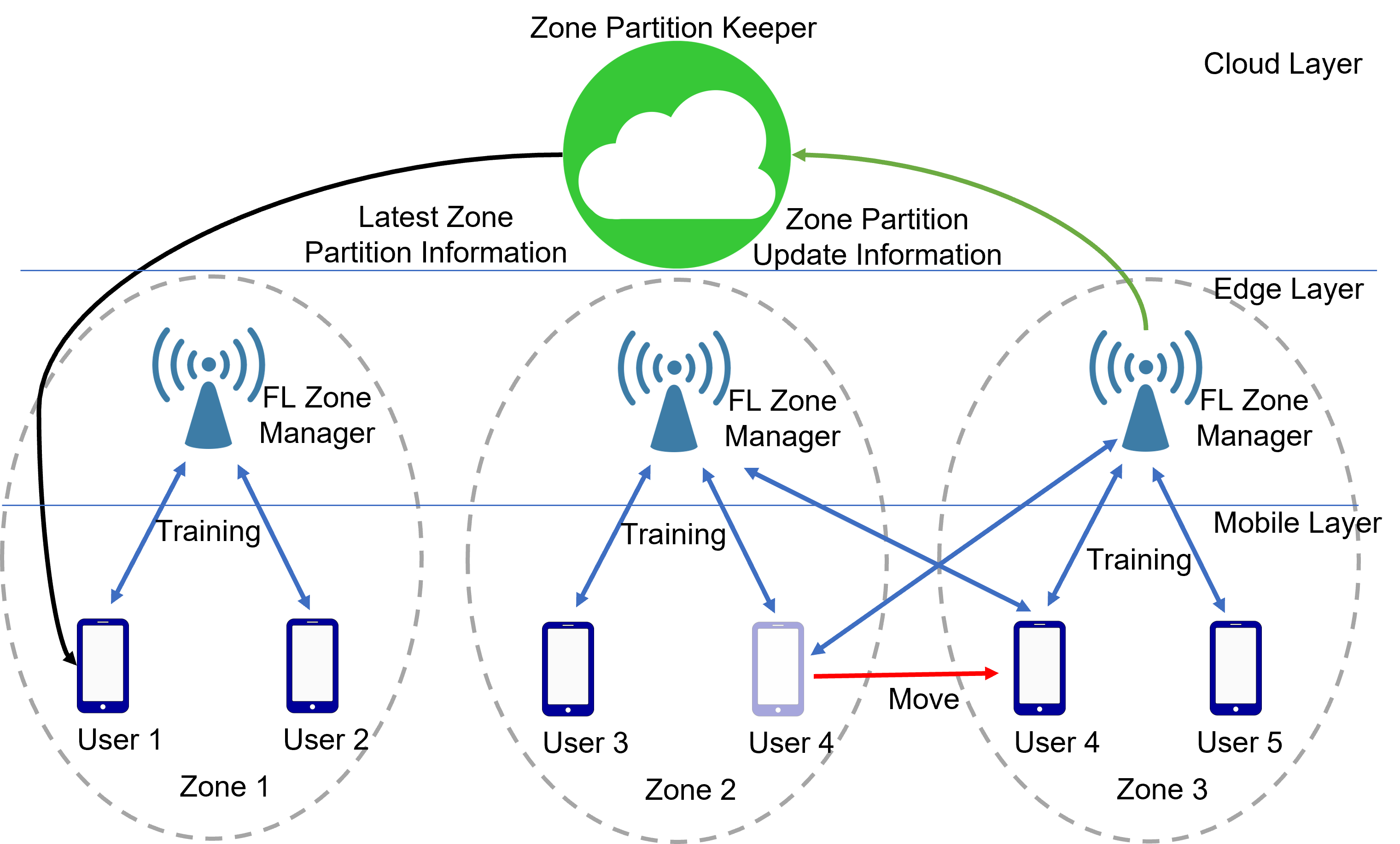}
\caption{ZoneFL Training Architecture.} \vspace{-15pt}
\label{fig:arch}
\end{figure}

\subsection{ZoneFL Training Overview} 

ZoneFL is designed to use a mobile-edge-cloud architecture, and its main goal is to train separate zone models (i.e., separate instances of the same base model) on mobile sensing data collected in each zone. Figure~\ref{fig:arch} shows the ZoneFL training and its high-level architecture. Each zone is managed by an FL Zone Manager at the edge, which maintains the latest models for its zone. A mobile is not tied to a single zone, but collects data from all the zones visited by the user and engages in training for each of these zones. For example, Figure~\ref{fig:arch} shows how \textit{User 4} moves from \textit{Zone 2} to \textit{Zone 3} and collects data in both zones. For each zone, mobiles that collected data in that zone train the zone models jointly with the edge zone manager. Mobile devices download the updated models their apps need from the edge managers when they need inference in a new zone (e.g., \textit{User 4} in \textit{Zone 3}). In this process, the FL Zone Manager at the edge does not know when and where the user was in a zone. It only knows the user has collected data in a zone, and need performance inference. Therefore, the potential privacy information that the edge can infer is very limited.
The cloud collaborates with the edge nodes to dynamically maintain the zone partition information for the entire space, as the geographical coordinates of the zones may change over time, but it is not involved in training. 
The mobile devices download the zone partition information and the identifiers of the edge managers from the cloud every time new zone configuration information is available.
The zone partition information is used to associate data with different zones, perform local training, and send the weights to the corresponding zone edge manager, which will aggregate the zone model.

The logical architecture of ZoneFL allows for the FL Zone Managers to be located in the cloud or at the edge. This decoupling of the software component from the hardware is useful until edge nodes will become widespread. Currently, the mobile-edge-cloud architecture is available only in certain major cities, etc. Nevertheless, the mobile-edge-cloud architecture provides better scalability than a mobile-cloud architecture because edge nodes in ZoneFL have a lower communication and computation load than the cloud server in traditional FL. Furthermore, the edge allows for faster interaction with the mobiles and for less bandwidth consumption in the network core. Finally, let us note that we assume only one edge node per zone. If there are multiple edge nodes in a zone, they can act as relays between the mobile devices and the node that runs the FL Zone Manager.


A major question in ZoneFL training is how to adapt the zone models to changes in user mobility behavior over time. We present two federated training algorithms that address this questions in different ways. First, Zone Merge Split (ZMS) dynamically adapts the zone partitions (i.e., the zone geographic coordinates) by either 1) merging two neighboring zones into a larger zone, whose model performs better than each of the individual zone models, or 2) splitting a larger zone back into previously merged smaller zones, whose individual models perform better than the model of the larger zone. Second, Zone Gradient Diffusion (ZGD) improves a zone model by aggregating contextual information derived from local gradients of neighboring zones. In ZGD, the zones do not change, but the user mobility behavior change is captured through the diffusion of information from neighboring zones. A self-attention mechanism is applied in ZGD to dynamically quantify the impact of each zone on its neighbors. Different deployments of ZoneFL may use either ZMS or ZGD or a combination of both based on trade-offs between model utility, scalability, and user mobility behavior. 

\subsection{Zone Merge and Split (ZMS)}
\label{subsec:zms}

ZMS is a dynamic zone management protocol that optimizes model utility across zones. In the following, we first formulate the merging and splitting of zones and show that the problem is NP-Hard. Based upon that, we approximate this NP-Hard problem using novel greedy algorithms for the two operations.
        
\textbf{Zone Merging}. Given a set of $N$ non-overlapping zones $Z = \{Z_i\}_{i \in [0, N]}$ and its complete set of possible combinations of zones $\Theta$, merging a zone $Z_i$ with its neighboring zones in $Z$ is to find the smallest set of non-overlapping and merged zones $\mathcal{Z} = \{\mathcal{Z}_j\}_{j \in [0, |\mathcal{Z}|]}$ where $\mathcal{Z} \in \Theta$, $|\mathcal{Z}|$ is the number of non-overlapping and merged zones in $\mathcal{Z}$, and $\cup_j \mathcal{Z}_j = \cup_i Z_i$ so that: \textbf{(1)} The model utility across merged zones $\sum_{\mathcal{Z}_j \in \mathcal{Z}}\calL(\btheta_j, \mathcal{Z}_j)$ is optimized (Eq. \ref{Cond1}); and \textbf{(2)} Every zone $Z_i$ achieves better model utility after merging (Eq. \ref{Cond2}). Note that $\calL(\btheta_j, \mathcal{Z}_j)$ is the loss function of a zone $\mathcal{Z}_j$ with the model parameters $\btheta_j$.
\begin{align}
    &\mathcal{Z}^*, \{\btheta_j^*\} = \argmin_{\{\btheta_j\}, \mathcal{Z} \in \Theta} \sum_{\mathcal{Z}_j \in \mathcal{Z}}\calL(\btheta_j, \mathcal{Z}_j) \label{Cond1} \\
    &\text{s.t. } \forall Z_i \in \mathcal{Z}_j: \calL (\btheta_j^*, Z_i) \leq \calL(\btheta^*_i, Z_i) \label{Cond2} 
\end{align}
where $\calL$ is a loss function, and the loss of a zone $\mathcal{Z}_j$ is an average loss over all the users' local data in that zone: $\calL(\btheta_j, \mathcal{Z}_j) = \frac{1}{|U_j|} \sum_{u \in U_j} \calL(\btheta_j, u)$ where $|U_j|$ is the number of users in the zone $\mathcal{Z}_j$.


\textbf{Zone Splitting}. Splitting a large zone into a set of smaller sub-zones is the reverse process of merging zones. Given a large zone $Z= \cup_{i \in [0, N]} Z_i$ formed by merging smaller sub-zones $\{Z_i\}_{i \in [0, N]}$ and $\boldsymbol{\Theta}$ is the set of all possible combinations of sub-zones $\{Z_i\}_{i \in [0, N]}$, splitting $Z$ is to find the smallest set of sub-zones $\mathcal{S} \in \boldsymbol{\Theta}$, such that: \textbf{(1)} The model utility across sub-zones is optimized; and \textbf{(2)} Every sub-zone $Z_i$ achieves better model utility after the zone splitting. 
\begin{equation}
\label{eq:split}
    \mathcal{S}^*, \{\btheta_j^*\} = \argmax_{\mathcal{S} \in \boldsymbol{\Theta}, \{\btheta_j^*\}} \cfrac{1}{|\mathcal{S}|} \sum_{\mathcal{Z}_j \in \mathcal{S}}[\calL(\btheta^*_Z, Z) - \calL(\btheta^*_j, \mathcal{Z}_j)]
\end{equation}

Eq.~\ref{eq:split} indicates $S^*$ is the (smallest) set of zones which has the maximal utility gain from the federated training of the original zone $Z$, i.e., $1 / |\mathcal{S}| \sum_{\mathcal{Z}_j \in \mathcal{S}}[\calL(\btheta^*_Z, Z) - \calL(\btheta^*_j, \mathcal{Z}_j)]$.

\textbf{Zone Merge and Split (ZMS) Algorithm.} ZMS is Np-hard problem, we propose ZMS, a greedy algorithm to dynamically adapt the zone models to changes in the user mobility behavior over time. 
In simple terms, ZMS merges two zones when the model performance of the merged zone is better than the performance of each of the models of the individual zones (i.e., to be merged). Each Zone Manager makes its own decisions regarding when to run the zone merging or zone splitting, as this decision depends on the conditions of each zone. For instance, the users in some zones may collect more data than the users in other zones, which may result in more frequent training. Also, the user behavior may change in some zones, while remaining similar in others. While running the zone merging and splitting in every training round may result in the best zone partitioning, such a solution results in too much overhead for both mobile users and Zone Managers. Therefore, we need to balance the trade-offs between zone partitioning efficiency and the computation and communication overhead.

Instead of checking all possible zone merges, ZMS randomly selects a zone $Z_i$ to check for possible zone merging at every round $t$. In the merging Algorithm~\ref{algo:merge}), ZMS merges $Z_i$ with its best neighboring zone $Z_n^*$, optimizing the zone merging objectives in Eqs. \ref{Cond1} and \ref{Cond2} (Alg. \ref{algo:merge}, Lines 2-7). 
The number of neighbors in line 2 is typically a small constant in practice, and lines 4-7 repeat over it. The additional round of training in line 5 trades computation cost for better performance improvement guarantee. It can be omitted, and $\btheta^{t+1}$ becomes $\btheta^{t}$ in lines 6 and 9. 
The best neighboring zone $Z_n^*$ is the zone that provides the maximal utility gain after the zone merging among all potential merges (Alg. \ref{algo:merge}, Line 9). To compute the utility gain, at the next training round $t+1$, we quantify the improvement of the loss in zones $Z_i$ and $Z_n$ using the zone models $\btheta^{t+1}_i$ and $\btheta^{t+1}_n$ trained respectively on $Z_i$ and $Z_n$ compared with using the zone model trained on the merged zone $Z_i \cup Z_n$. 

The zone models are trained and validated in the background by the phones in their respective zones. Mobile phones retain a small validation dataset to validate the zone models, and send the validation results to their zone manager to be used in merge decisions. Thus, these operations do not incur latency during merges. The only operation that needs to be done specifically for a merge is the validation of the model over the two zones. To reduce the overhead, the zone manager to select only a percentage $p$ of the phones in its zone to perform training and validation in this case. 

Merging in ZMS also handles the case when the original zones set during bootstrapping do not have enough data for adequate training. In this situation, ZMS will merge such zones with neighboring zones, therefore improving performance. 

\begin{algorithm}[t]
\caption{Zone Merging Algorithm}\label{algo:merge}
\footnotesize
\textbf{Input}: Zone $Z_i$
\begin{algorithmic}[1]
 \STATE $C \leftarrow \emptyset$ \# initialize a list of zone merging candidates
 \STATE $\calN \leftarrow \textbf{\textit{getNeighbors}}(Z_i)$ \# get neighboring zones of $Z_i$
  \FOR{each neighboring zone $Z_n \in \calN$}
    \STATE $\btheta^t_{in} \leftarrow (\btheta^t_i + \btheta^t_n)/2$ \# average of two zone models 
    \STATE $\btheta^{t+1}_{in} \leftarrow \argmin_{\btheta_{in}} \calL(\btheta^t_{in}, Z_i \cup Z_n)$ \# Eq. \ref{Cond1}
    \IF{$\calL(\btheta^{t+1}_{in}, Z_i) < \calL(\btheta^{t+1}_{i}, Z_i)$ and $\calL(\btheta^{t+1}_{in}, Z_n) < \calL(\btheta^{t+1}_{n}, Z_n)$ \# satisfying Eq. \ref{Cond2}}
        \STATE $C \leftarrow C \cup Z_n$ \# add $Z_n$ into a list of candidates
    \ENDIF
  \ENDFOR

\IF{$C \neq \emptyset$}
        \STATE $Z_n^* \leftarrow \arg\max_{Z_n \in C} \big[\calL\left(\btheta^{t+1}_{in}, Z_i\right) - \calL(\btheta^{t+1}_i, Z_i) \big] + \big[\calL\left(\btheta^{t+1}_{in}, Z_n\right) - \calL(\btheta^{t+1}_n, Z_n)\big]$ \# get the best neighboring zone
        \STATE $\textbf{\textit{Merge}}(Z_i, Z_n^*)$
\ENDIF  
\end{algorithmic}  
\end{algorithm}

\begin{algorithm}[t]
\caption{Zone Splitting Algorithm}
\label{algo:split}
\footnotesize
\textbf{Input}: Zone $\mathcal{Z}_j = \cup_{i} Z_i$, level $l$
\begin{algorithmic}[1]
 \STATE $C \leftarrow \textbf{\textit{getCandidates}}(\mathcal{Z}_j, l)$
  \FOR{each zone $Z_c \in \textrm{top-}k(C)$}
    \STATE $\btheta^{t+1}_{c} \leftarrow \argmin_{\btheta_c} \calL(\btheta^t_{j}, Z_c)$
    \IF{$\calL(\btheta^{t+1}_{c}, Z_c) < \calL(\btheta^{t+1}_{j}, Z_c)$}
        \STATE $\textbf{\textit{split}}(\mathcal{Z}_j, Z_c)$ \# split the sub-zone $Z_c$ from the merged zone $\mathcal{Z}_j$
        \STATE \textbf{break}
    \ENDIF
  \ENDFOR

  \textbf{Function} \textbf{\textit{getCandidates}}($\mathcal{Z}_j, l$):

  \STATE $C \leftarrow \emptyset$ \# initialize a list of worst sub-zones

  \FOR{$Z_c \in \text{\textbf{\textit{subZones}}}(\mathcal{Z}_j, l)$}
    \IF{$\calL(\btheta^{t}_j, Z_c) > \calL(\btheta^{t}_j, \mathcal{Z}_j)$}
        \STATE $C \leftarrow C \cup Z_c$
    \ENDIF
  \ENDFOR
  \RETURN $sorted(C)$ \# descending $\calL(\btheta^{t}_j, Z_c)$
  
\end{algorithmic} 
\end{algorithm}


ZMS repeats this zone merging process across federated training rounds to create a set of merged zones, denoted $\mathcal{Z}$. However, over time, in response to user mobility behavior changes, some of the merged zones may need to be split. 

\begin{wrapfigure}{l}{0.28\textwidth}
  \begin{center}
\includegraphics[width=0.28\textwidth]{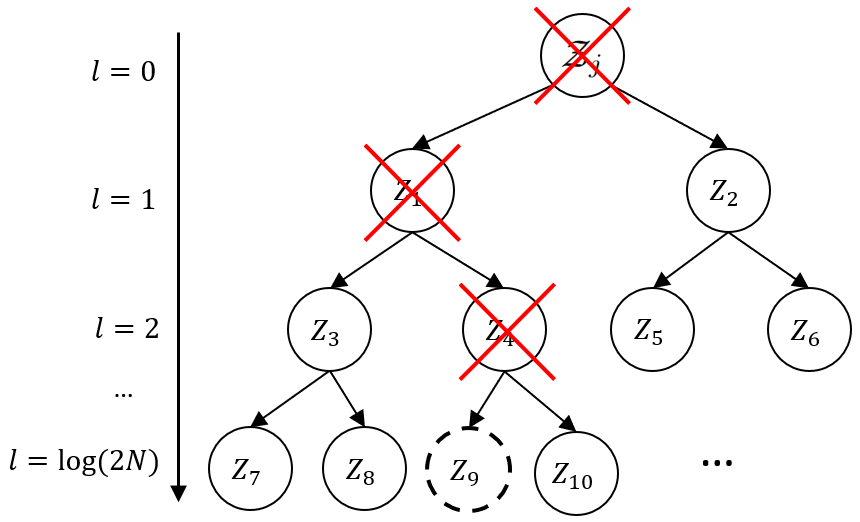} \vspace{-20pt}
  \end{center}
  \caption{Binary Tree and Zone Splitting}
  \label{fig:tree}
\end{wrapfigure}

The key idea of zone splitting is to identify the zone that performs worst in terms of model utility and split it from an original merged zone so that the zones after splitting perform better than the original zone. Specifically, ZMS recursively split sub-zones of a merged zone, which have the worst model utility, such that the splitting optimizes model utility across all sub-zones. 
Each of the merged zones $\mathcal{Z}_j \in \mathcal{Z}$ is a set of sub-zones $\{Z_i\}_{i \in [1, N]}$ represented by a binary tree of zone merging history, as illustrated in Figure \ref{fig:tree}. Each internal node in the tree represents a merged zone from its two sub-zones (child nodes). Each leaf node is an indivisible zone. 
\indent At each training round, ZMS randomly selects a binary tree representing a merged zone $\mathcal{Z}_j$ to check for a potential zone splitting. 
ZMS considers all the internal nodes up to level $l$ as potential sub-zones to split. 
For instance, if $l = 2$ in Fig. \ref{fig:tree}, we consider $\{Z_i\}_{i \in [1, 6]}$ as candidates for the zone splitting (Alg. \ref{algo:split}, Line 1). 
We select top-$k$ sub-zones having inferior model utility (i.e., higher losses compared with the merged zone $\mathcal{Z}_j$) (Alg. \ref{algo:split}, Lines 7-15). If a candidate zone $Z_c$ trained independently achieves better model utility, i.e., $\calL(\btheta^{t+1}_{c}, Z_c) < \calL(\btheta^{t+1}_{j}, Z_c)$ where $\btheta^{t+1}_{c}$ is the zone model trained on $Z_c$ and $\btheta^{t+1}_{j}$ is the model trained on the merged zone $\mathcal{Z}_j$ (Alg. \ref{algo:split}, Line 4), then ZMS splits $Z_c$ from the merged zone $\mathcal{Z}_j$ (Alg. \ref{algo:split}, Line 5). In a training round, ZMS permits at most one zone splitting (Alg. \ref{algo:split}, Line 6) to minimize the overhead and avoid distributed consistency problems. 
All ancestor nodes of $Z_c$ are removed, creating a set of new merged-zones and their associated binary trees. For instance, in Fig. \ref{fig:tree}, if we split zone $Z_9$, we create a set of new merged zones, including zones $Z_3 = Z_7 \cup Z_8$, $Z_2 = Z_5 \cup Z_6$, $Z_9$, and $Z_{10}$.
By doing this, we focus on keeping the best merges after a zone splitting; thus approximating the zone splitting objective (Eq. \ref{eq:split}) without affecting the zone merging objectives (Eqs. \ref{Cond1} and \ref{Cond2}). The training and validation at the phones for split is done in a similar way with the ones for merge.


\begin{algorithm}[t] 
\caption{Zone Gradient Diffusion with Self-Attention}\label{algo:Diff-ZFL}
\footnotesize
\textbf{Input}: Zone $Z_i$
\begin{algorithmic}[1]
 \STATE $\calN_i \leftarrow \textbf{\textit{getNeighbors}}(Z_i)$
  \FOR{$Z_n \in \calN_i$}
    \STATE $e_{in} \leftarrow \sigma \big(\grad(\btheta_i^{t}, Z_i) \bullet \grad(\btheta_i^{t}, Z_n) \big)$ \# where ``$\bullet$'' is an inner product
    \ENDFOR
    \STATE $\forall Z_n \in \calN_i: \beta_{in} \leftarrow \cfrac{\exp(e_{in})}{\sum_{Z_j \in \calN_i} \exp(e_{ij})}$ \# computing coefficients
    \STATE $\btheta_i^{t+1} \leftarrow \btheta_i^{t} + \grad(\btheta_i^{t}, Z_i) + \sum_{Z_n \in \calN_i} \beta_{in}\grad(\btheta_i^{t}, Z_n)$ \# aggregating gradients from neighboring zones
\end{algorithmic} 
\end{algorithm} 

\subsection{Zone Gradient Diffusion (ZGD)}


In addition to ZMS, we propose ZGD, an algorithm that keeps the zones fixed but adapts the model by aggregating contextual information derived from local gradients of neighboring zones (Alg. \ref{algo:Diff-ZFL}). We found that contextual information captures changes in mobility patterns and significantly improves the utility of zone models. In ZGD, at round $t$, the neighboring zones $Z_n$ of a zone $Z_i$ derive their local gradients using the model parameters $\boldsymbol{\theta}_i^{t}$ from the zone $Z_i$ by using local data $D_u$ from their users $u$, as follows: $\grad(\boldsymbol{\theta}_i^{t}, Z_n) = 1/|U_n| \sum_{u \in U_n} \grad(\boldsymbol{\theta}_i^{t}, D_u)$. Note that users $u$ compute the gradients $\grad(\boldsymbol{\theta}_i^{t}, D_u)$ and send the gradients to the zone manager $Z_n$ for data privacy protection.

Intuitively, the more similar the gradients of a zone ($\grad(\boldsymbol{\theta}_i^{t} Z_i)$) are with the gradients of a neighboring zone ($\grad(\boldsymbol{\theta}_i^{t}, Z_n)$), the higher the impact of the neighboring zone $Z_n$ on $Z_i$ will be. We quantify this impact through self-attention coefficients $\beta_{in}$ by normalizing the inner product of the local gradients of the zone $Z_i$ and its neighboring zones $Z_n \in \mathcal{N}_i$:
\begin{equation}
\forall Z_n \in \calN_i: \beta_{in} \leftarrow \cfrac{\exp(e_{in})}{\sum_{Z_j \in \calN_i} \exp(e_{ij})}
\end{equation}
where $e_{in} = \sigma \big(\grad(\btheta_i^{t}, Z_i) \bullet \grad(\btheta_i^{t}, Z_n) \big)$, $\sigma$ is the sigmoid function, and ``$\bullet$'' is an inner product.

Finally, we aggregate the gradients from neighboring zones to update the zone model $\boldsymbol{\theta}_i^{t}$ at round $t$:
\begin{equation}
    \boldsymbol{\theta}_i^{t+1} \leftarrow \boldsymbol{\theta}_i^{t} + \grad(\boldsymbol{\theta}_i^{t}, Z_i) + \sum_{Z_n \in \mathcal{N}_i} \beta_{in} \grad(\boldsymbol{\theta}_i^{t}, Z_n)
\label{eq:zgd}
\end{equation}

By doing so, ZGD updates the zone models to diffuse contextual information from one zone to all the remaining zones across training rounds. This operation significantly enriches the information used to optimize zone models in ZoneFL, compared with existing FL algorithms.

\section{System Design and Implementation}
\label{sec:arch}


\subsection{System Architecture}

\begin{figure}[t]
\centering
\includegraphics[width=1\linewidth,scale=1]{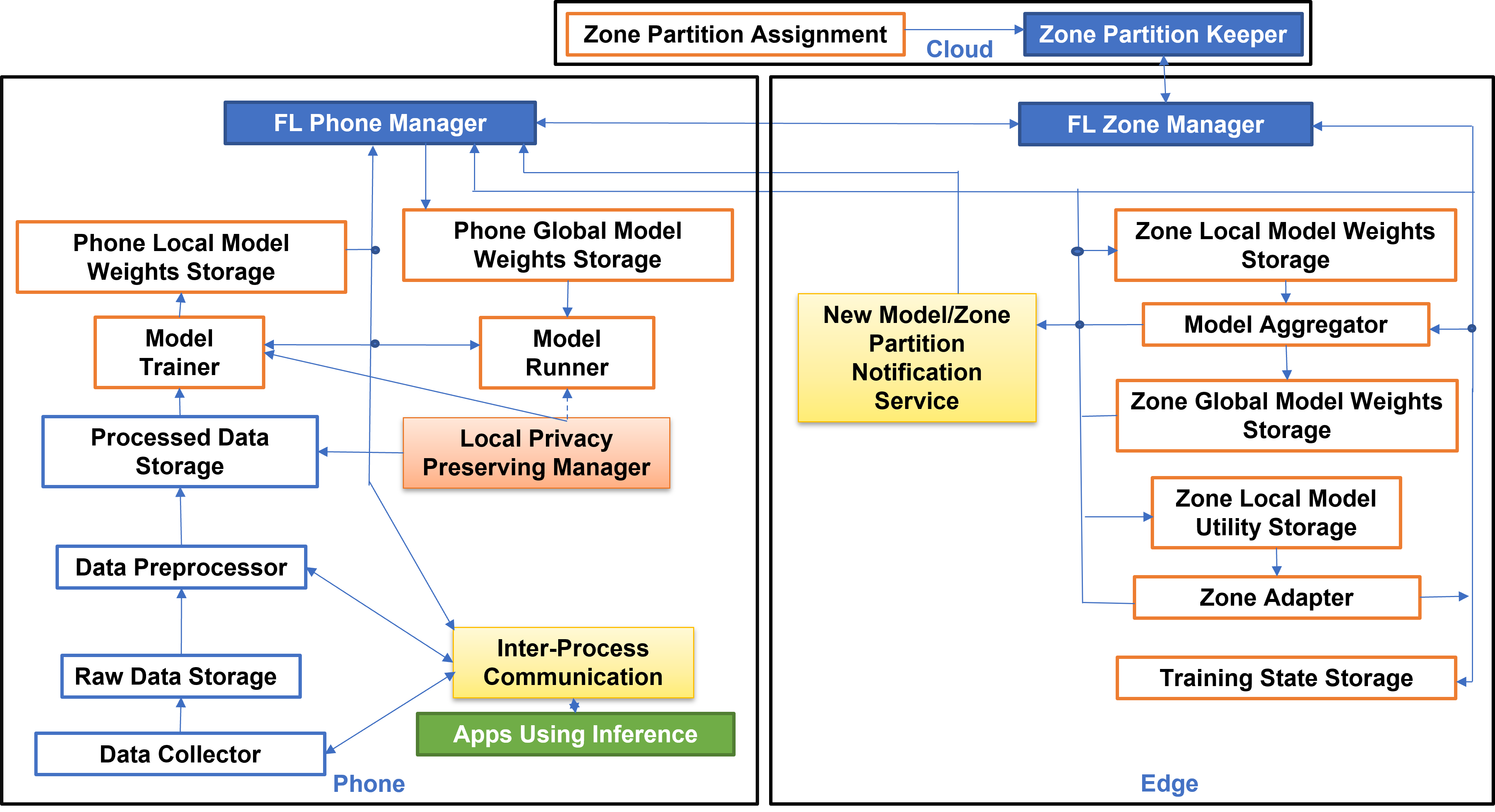}
\caption{System Architecture.}
\label{fig:soft}
\end{figure}

The ZoneFL architecture has three main components, as shown in Figure~\ref{fig:soft}: \emph{(1)} \emph{FL Phone Manager} coordinates the ZoneFL activities on the phone; \emph{(2)} \emph{FL Zone Manager} coordinates the ZoneFL activities at the edge; and \emph{(3)} \emph{Zone Partition Keeper} maintains and provisions the latest zone partition information in the cloud. The edge software components of the architecture can be mapped either to edge nodes or to servers in the cloud. For example, some FL Zone Managers could be deployed at the edge nodes where edge is available, while others can be hosted in the cloud where edge is not available yet. The FL Zone Manager can be migrated between the cloud and the edge nodes.

The software components work together to support the six phases of ZoneFL: data collection and preprocessing, privacy protection, model training and aggregation, mobile apps using models for inference, zone partition maintenance, and zone partition adaptation to user mobility changes. The first four phases follow traditional FL. The Data Collector stores the sensed data in the Raw Data Storage and informs the FL Phone Manager each time new data is added to the Raw Data Storage. The FL Phone Manager decides invokes the model-specific Data Processors and stores the data in the Processed Data Storage. The Local Privacy Preserving Manager uses differential privacy techniques to further preserve user privacy. The Model Trainer performs local training on the phone, and the Model Aggregator aggregates the gradients at the edge.
A Publish-Subscribe edge service, New Model/Zone Partition Notification Service, allows the phone to receive asynchronous notifications when a new zone model is available. When an app needs inference from a model, it sends a request to the FL Phone Manager using the OS IPC mechanisms. In response, the FL Phone Manager generates the input for the inference from the data stored in the Processed Data Storage, and then it invokes the Model Runner with this input. The Model Runner sends the result to the App using IPC. In the following, we explain the two phases that are specific to ZoneFL.

\textbf{Zone Partition Maintenance.} The Zone Partition Keeper maintains the latest zone partition information in the system, which is represented as a graph. Each non-overlapping zone is a vertex, and each edge connects two neighboring zones. The initial zone partition information is bootstrapped by the administrator of the system based on administrative divisions of a region. The Zone Partition Keeper is also responsible for maintaining information about the identity (e.g., IP addresses) of the FL Zone Managers at the edge. 

Initially, a phone receives the zone partition information from the Zone Partition Keeper. Then, it maps its data to different zones, based on the geographic locations where the data were collected. This determines the list of zones to which the phone subscribes for training. The phone communicates with the FL Zone Managers of these zones to jointly train the zone models. For inference, a phone may use a zone model even if the phone did not  participate in the training of the given zone. This allows new users to quickly benefit from ZoneFL.

\textbf{Zone Partition Adaptation.} 
The Zone Adapter of each edge node is responsible for dynamic adaptation of zone partitions in ZMS. In order to perform merge and split, as described in Section~\ref{subsec:zms}, the system needs to perform zone level model validation. This operation is done through the cooperation of the phones and the edge manager. The FL Zone Manager maintains a Zone Local Model Utility Storage for phones to report the model utility computed on their validation datasets, and periodically aggregates the validation results. This process involves additional communication between phones and the FL Zone Manager, but it mitigates potential privacy issues, since data never leaves the phone. 

\subsection{ZoneFL Prototype Implementation}
\label{imp}

We implemented an end-to-end ZoneFL prototype on Android phones and AWS cloud. This prototype, with ZMS for dynamic adaptation, was used in our field study, described in Section~\ref{sec:exp}.
AWS offers AWS Local Zones as its edge computing service. However, it is not available yet in the area of our field study, and therefore the edge components of ZoneFL are deployed in the AWS cloud. We chose Deep Learning for Java (DL4J) as the underlying framework for DL-related operations, because it is a mature framework that supports model training on Android devices. 

\textbf{Deployment and Operation Scripts.} The system administrator prepares the initial zone partition information as a geojson file, which defines the zones' geometry as polygon coordinates. 
We implement Python scripts to deploy and operate the system. These deployment script reads the geojson file provided by the system administrator to create an independent FL Zone Manager for each zone in AWS. The operation scrips are used to collect performance and reliability data. 

\textbf{FL Zone Manager.}  The core computing components of the FL Zone Manager are implemented and deployed as AWS Lambda functions~\cite{lambda} for low overhead and fast start time. We create a REST API to relay clients' requests to participate in the FL training to the Lambda function that handles these requests. We also use the AWS EventBridge to define rules to trigger and filter events for Lambda functions. For model storage, model utility storage, validation datasets, and configuration files, we use AWS S3. To store data that is accessed frequently, such as training round states and model states, we use AWS DynamoDB.  AWS SNS is utilized in conjunction with the Google FCM to notify clients when newly trained models are ready. Most FL Zone Manager components interact only with components within their zone. The only exception is the Zone Adapter, which communicates with its counterparts in neighboring zones to implement ZMS. 
As public cloud providers are racing to deploy edge computing infrastructure~\cite{aws-local-zones,azure-edge-zones}, we expect these cloud services or their edge-based variants will soon be available at the edge. 

\textbf{Zone Partition Keeper.} We use an AWS S3 bucket as the Zone Partition Keeper of all zones. This is the only shared AWS resource in the system. All the other AWS resources are independent among different zones. In this way, once edge computing becomes more widespread, the FL Zone Manager can be migrated from the cloud to the edge. The latest zone partition information is made available to phones for download. The previous partition information is also stored for the Zone Adapter to help with the split operation in ZMS.  

\textbf{Android implementation.} The Android phone implementation consists of three apps: FL Phone Manager, Data Collector, and Testing App (used to test model inference). The Data Collector was implemented starting from ExtraSensory~\cite{extrasensory}. This app collects heart rate (HR) sensing data from a Polar HR tracking wrist band~\cite{polar}, which connects to the phone over Bluetooth. In the FL Phone Manager, the Data Preprocessor uses the geojson file with Android Google Map API to check the zone to where each data point belongs to.  Then, the Data Preprocessor generates the model input for training.
The Model Trainer is implemented with the Android native \emph{AsyncTask} class to ensure the trainer is not terminated by Android, even when the app is idle. The Model Trainer communicates with the FL Zone Manager of each zone to train the models sequentially. Model inference is implemented as a background service with Android Interface Definition Language (AIDL), and it gets inference requests from the Testing App. This app uses \emph{AidlConnection} to interface with the FL Phone Manager for the inference results.
\section{Evaluation}
\label{sec:exp}

The evaluation presents results for both model utility and system performance. The model utility experiments have two goals: \textbf{(i)} Compare the performance of ZoneFL with Global FL (i.e., traditional FL trained with all users globally); \textbf{(ii)} Quantify the benefit of ZGD and ZMS. The system experiments have four goals: \textbf{(i)} Demonstrate the feasibility of ZoneFL on smart phones; \textbf{(ii)} Investigate ZoneFL scalability; and \textbf{(iii)} Quantify the ZoneFL phone training time overhead.

\subsection{Datasets, Models, and Metrics}
We use two datasets collected in the wild to evaluate two types of ZoneFL models: \textbf{(1)} A human activity recognition dataset~\cite{hu2021flsys}; and \textbf{(2)} A heart rate dataset~\cite{fitrec}. We choose these two datasets because we observe the advantages of ZoneFL with these two real-world mobile sensing applications we have data. The attributes, other than zone, that affect the prediction are handled by the model design.

\textbf{Human Activity Recognition (HAR).} The dataset has data from 51 users, moving in a region larger than 20,000 $km^2$. 
Each user provided mobile accelerometer data, GPS coordinates, and labeled their daily activities on their personal Android phones. The labels used in the experiments are ``Walking,'' ``Sitting,'' ``In Car,'' ``Cycling,'' and ``Running.''
In the experiments, we start with 9 non-overlapping zones over the region covered by the dataset, based on GPS coordinates. The zones are diverse and include a university campus zone, several suburban residential zones, a riverside urban zone, a metro zone, etc.  On average, each user have 1,995 data samples for each zone. 
The preprocessing and the CNN-based model architecture follow the work associated with the dataset~\cite{hu2021flsys}.
For this classification task, we use accuracy as the main metric for model performance.

\textbf{Heart Rate Prediction (HRP).} The dataset contains 167,373 workout records for 956 users in 33 countries. The data collected by the users using their mobile/wearable devices include multiple sources of sequential sensor data such as heart rate, speed, GPS, sport type, user gender, and weather conditions. We filter the data such that users with at least 10 workouts are included in training and inference (4:1 split). We exclude users having less than 10 workouts because those data points are not significant. To evaluate ZoneFL, we assign the initial zones of each country to its principal (largest) administrative divisions so that we can have a manageable number of zones. Among the countries, we select the top 6 countries having at least 10 zones with a reasonable number of average data samples per zone to effectively assess ZoneFL's performance.
We use an LSTM-based model~\cite{fitrec} to predict the heart rate given input features consisting of altitude, distance, and time elapsed (or speed) of the workouts. For this prediction task, we use the root mean squared error (RMSE) metric. We only use HRP to evaluate zone dynamic adaptation because HRP dataset has sufficient number of zones and users.

\subsection{Model Utility Results}

\textbf{ZoneFL vs. Global FL.} Table~\ref{global} shows the performance comparison between ZoneFL and Global FL. In this experiment, ZoneFL works only with the zones defined at the beginning of the experiment (called Static ZoneFL), without employing ZMS or ZGD to adapt the models to the user mobility behavior over time. Thus, it provides a lower bound on ZoneFL's performance, which is expected to improve with ZMS and ZGD. Global FL trains with all the users in the datasets. Zone FL trains a different model for each zone in the respective dataset. Some users have data and participate in training in more than one zone. The metrics are computed per user in the test data set and then averaged. ZoneFL models outperform the Global FL models by 6.67\% for HAR and by 6.74\% for HRP. This performance gain is significant given that it is very challenging for DL models to achieve 1\% improvement in HAR and HRP tasks as illustrated in recent studies~\cite{chen2021deep, ni2019modeling}. As shown next, we observe further improvement with the dynamic adaptation algorithms.

\begin{table}[t!]
\centering
\caption{ZoneFL vs. Global FL}
    \vspace{-5pt}
\resizebox{0.49\textwidth}{!}{%
\begin{tabular}{|l|l|l|l|l|}
\hline
Application & Metrics & Global FL & Static ZoneFL & Improvement Gain       \\ \hline
HAR & Accuracy (\%)    & 65.27 & 69.63 & 6.67\% \\ \hline
HRP & RMSE      & 21.20  & 19.86 & 6.74\% \\ \hline
\end{tabular}}  \vspace{-10pt}
\label{global}
\end{table}

\begin{figure}[t]
\centering
\includegraphics[width=0.85\linewidth,scale=1]{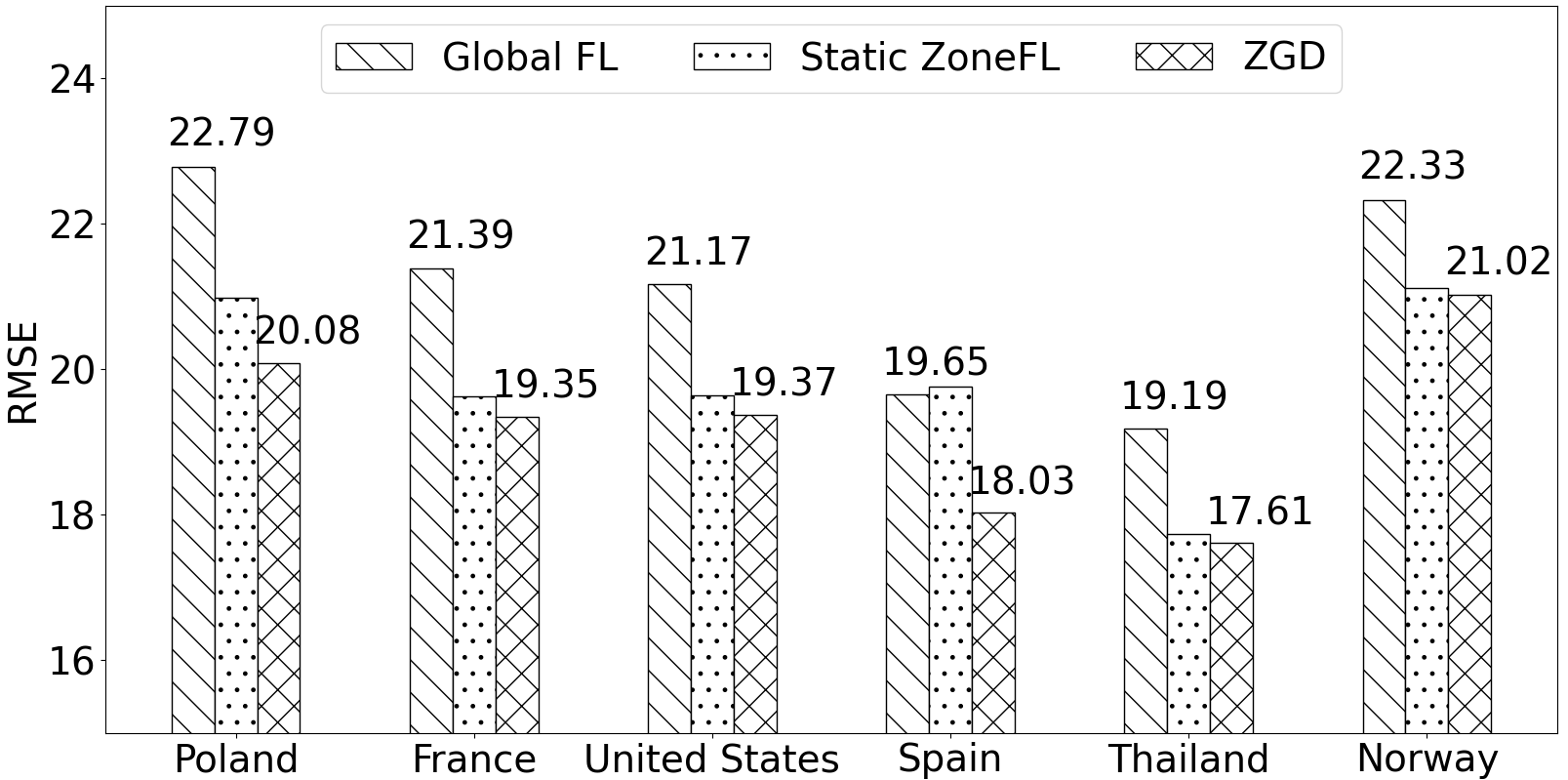}
\vspace{-5pt}
\caption{Simulation Results of Global FL and ZoneFL Algorithms.} \vspace{5pt}
\label{fig:zgd}
\end{figure}

\textbf{ZGD Performance.} Although ZGD and ZMS adapt zone models to user mobility behavior changes, they serve slightly different purposes. Hence, we present the performance of ZGD and ZMS separately. ZGD is designed to work with fixed zones that have enough data for training. ZMS is designed to adapt the zone partitions until all of them achieve reasonable model performance. In practical terms, ZMS is generally used for the beginning rounds of ZoneFL, while ZGD is used once the zone model performance is relatively stable. For both algorithms, we show just the results for HRP because its dataset is more suitable for dynamic adaptation by having more zones.

Figure~\ref{fig:zgd} shows the performance of ZGD for the top-6 countries in the HRP dataset. ZoneFL with ZGD performs better than Static ZoneFL for each country, and it clearly outperforms Global FL (by as much as 11.89\% for Poland). We also observe that Static ZoneFL performs better than Global FL for 5 countries, and slightly worse for one country. The reason for the worse performance for Spain is that the static zones do not capture well the changes in user mobility behavior. ZoneFL with ZGD is able to alleviate this problem and result in better performance than Global FL. 

\textbf{ZMS Performance.} Table~\ref{hrp_MS} shows the average model performance improvement for (zone) merge and split in HRP. In merge, the improvement gain is calculated as follows: $\frac{L_1 + L_2}{2} - L_{12}$, where $L_1$ and $L_2$ are RMSE losses evaluated on the two constituent zones, and $L_{12}$ is RMSE loss computed on the merged zone. The reverse formula is used for splitting a larger zone in two sub-zones. The results demonstrate that ZMS can significantly improve the model performance. On average, 4 merges and 3 splits occur every 100 rounds of training, which shows that dynamic adaptation needs to happen about once a month in a scenario where users train once a day.

\begin{table}[t!]
\centering
\caption{ZMS Improvement}
    \vspace{-5pt}
\resizebox{0.45\textwidth}{!}{%
\begin{tabular}{|l|l|l|l|l|}
\hline
  & \shortstack{Before \\(RMSE)} &  \shortstack{After\\  (RMSE)} &\shortstack{Improvement Gain \\ (\%)  Mean / SD} & \shortstack{Occurrence  Per \\ 100 Rounds}     \\ \hline
Merge & 23.79  & 21.44 & 9.87 / 3.11    & 4 \\ \hline
Split &  23.04  &   20.71   &     11.10 / 3.63    & 3 \\ \hline
\end{tabular}}  
\label{hrp_MS}
\end{table}




\subsection{System Results}

To showcase the feasibility and advantages of ZoneFL over Global FL in a real-life deployment, we conducted an HRP field study with 63 users for 4 months. Along with smart phone sensor data such as accelerometer, gyroscope, etc., the users were tasked to collect heart rate data from a Bluetooth-connected heart rate tracking wrist band for their daily activities. The region of the field study is larger than 20,000 $km^2$, and it was originally divided in 9 zones. The study ran the prototype of ZoneFL with ZMS, described in Section~\ref{imp}. In the field study the ZMS split operation is performed for only one level ($l = 1$, Section~\ref{subsec:zms}). The prototype worked reliably throughout the duration of the field study. Next, we present experimental results for our prototype.


\subsubsection{ZoneFL Feasibility on Smart Phones}
\label{feasibility}

We benchmarked ZoneFL with HAR and HRP on Android phones using a testing app to evaluate training and inference performance. We also assessed the resource consumption on the phones, with different specs (Nexus 6P and Google Pixels 3). The results demonstrate the on-device feasibility of ZoneFL, even for the Nexus 6P phone, unveiled in 2015 and running Android 7. Since ZoneFL works well on such a low-end phone, we expect ZoneFL to work well on most of today's phones.

\begin{table}[t!]
\centering
\caption{Training on Phones: Resource Consumption and Latency} 
    \vspace{-5pt}
\resizebox{0.5\textwidth}{!}{%
\begin{tabular}{|l|l|l|l|l|l|l|}
\hline
Application & Phone         
                & \shortstack{ Max \\RAM\\ Usage \\ (MB) } & \shortstack{ Foreground \\ Training \\ Time \\ Mean/SD\\ (min) } & \shortstack{ Background \\ Training \\ Time  \\ Mean/SD \\(min) }
                 & \shortstack{Battery \\ Consumption \\ per Round \\ (mAh)}  & \shortstack{ Number of \\Training \\ Rounds \\for Full\\ Battery }   \\ \hline
\multirow{2}{*}{HAR} 
& Nexus 6P     &232  & 15.21/2.89   & 59.99/4.06       & 53.86  & 64   \\ \cline{2-7}
& Google Pixel 3      &228  & 2.13/0.24  & 9.32/0.09    & 9.91  & 294   \\ \hline
\multirow{2}{*}{HRP} 
& Nexus 6P       &266  & 3.09/0.39  & 10.97/1.08    & 33.18  & 104   \\ \cline{2-7}
& Google Pixel 3      &230  & 0.40/0.10  & 5.07/0.37   & 4.66  & 625   \\ \hline

\end{tabular}} 
\label{training} 
\end{table} 

\textbf{Training Performance.} Table~\ref{training} shows the ZoneFL training time and  resource consumption on the phones. The training time is recorded by training 1995 samples and 86 samples (i.e., the average numbers of samples per zone per user) in 5 epochs for HAR and HRP. Foreground training (screen turned on) provides a lower bound for the training time by using the full single core capacity. In reality, we expect training to be done in the background, while the phone is being charged. We take 10 measurements for each benchmark and report the mean and standard deviation since other apps or system processes working in background may interfere with the training.

Training for one round is fast on the phones. The foreground training time on Pixel 3 is just 2.13 min for HAR, and 0.4 min for HRP. The background training time is also good for any practical situation. The background training time is notably longer compared with foreground training, since Android attempts to balance computation with battery savings.

The results also show training is feasible in terms of resource consumption. The maximum RAM usage of the app is less than 266MB, and modern phones are equipped with sufficient RAM to handle it. The phones could easily perform hundreds of rounds of training on a fully charged battery. It is worth noting that, typically, one round of training per day is enough, as the users need enough time to collect new data.

\begin{table}[t!]
\centering
\caption{Inference on Phones: Resource Consumption and Latency}
    \vspace{-5pt}
\resizebox{0.5\textwidth}{!}{%
\begin{tabular}{|l|l|l|l|l|l|l|}
\hline
Application & Phone          
                & \shortstack{ Max \\RAM\\ Usage \\ (MB) } & \shortstack{ Foreground \\ Inference \\ Time \\ Mean/SD\\ (millisecond) } & \shortstack{ Background \\ Inference \\ Time \\ Mean/SD \\(millisecond) }
                 & \shortstack{Battery \\ Consumption \\ per \\prediction \\ ($\mu$Ah)}  & \shortstack{  Millions of \\inferences\\ for \\Full \\Battery }   \\ \hline
\multirow{2}{*}{HAR} 
& Nexus 6P       &161  & 54.65/16.36  & 1963.04/1540.29      & 4.49  & 0.77   \\ \cline{2-7} 
& Google Pixel 3        &177  & 36.59/6.43  & 99.60/33.69    & 1.94  & 1.50   \\ \hline

\multirow{2}{*}{HRP} 
& Nexus 6P       &232  & 528.93/53.53  & 1809.71/700.96      & 45.47  & 0.08   \\ \cline{2-7}
& Google Pixel 3        &229  & 167.71/6.83  & 669.88/112.01   & 5.74  & 0.51   \\ \hline

\end{tabular}} 
\label{inference}
\end{table}

\textbf{Inference Performance.} The results in Table~\ref{inference} demonstrate that ZoneFL can be used efficiently by third-party apps working in real-time. 
The inference time is measured within the third party testing app. Let us note that the inference is performed locally by the FL Phone Manager, without any network communication. Thus, the measured time consists of the inference computation time and the inter-process communication time. We continuously perform predictions/classifications for 30 minutes and report the average values. The inference time for the two scenarios on the third-party app, foreground and background, follows a similar trend as training. 

\subsubsection{Scalability} 

ZoneFL utilizes multiple FL Zone Managers to receive and aggregate the gradients from the users. Compared with a single server in Global FL, the communication and computation load in ZoneFL is distributed among multiple zone servers. Considering a user may send gradients to multiple zone servers, Table~\ref{scale} computes the average ZoneFL server load savings based on the user percentage distribution over the number of zones in Figure~\ref{fig:overhead}. The results demonstrate ZoneFL scales better than Global FL because the server load is 34.98\% to 37.26\% of the one in Global FL.

\begin{table}[t!]
\centering
\caption{Server Load in ZoneFL over Global FL }
    \vspace{-5pt}
\resizebox{0.3\textwidth}{!}{%
\begin{tabular}{|l|l|l|}
\hline
Application & HAR & HRP             \\ \hline
ZoneFL server load &37.26\% & 34.98\%    \\ \hline
\end{tabular}} \vspace{5pt}
\label{scale}
\end{table}

\subsubsection{ZMS Performance in the Field Study} 
Table~\ref{merge} depicts zone merge time and model utility gains in the field study. At the end of the field study, the number of the zones was changed from 9 to 7 after several merges and splits. In ZMS, a merge occurs when the merged model performs better than both individual zone models. The highest model utility gain observed is to improve RMSE from 44.53 to 10.84. This is because the original zones did not have enough users and data. We also observed two splits happened during the filed study. The highest RMSE improvement for split is from 16.38 to 11.20. These observations showcase the ZMS improvements in our ZoneFL prototype deployed in real-life.


\begin{table}[t!]
\centering
\caption{ZMS in The Field Study } \vspace{-5pt}
\resizebox{0.4\textwidth}{!}{%
\begin{tabular}{|l|l|l|l|}
\hline
Merge Time  &  \shortstack{ Zone X/\\RMSE }& \shortstack{Zone Y/\\RMSE} & \shortstack{Merged\\ Zone RMSE }             \\ \hline
2022-04-09 13:57       &A/13.96  &B/18.40  & 12.56 \\ \hline
2022-05-29 12:53        &C/44.53  &D/11.86  & 10.84   \\ \hline

2022-06-05 13:07      &E/18.48  &A/15.28  & 13.30  \\ \hline
2022-07-29 21:56    &F/17.40  & G/39.23  & 14.78  \\ \hline
\end{tabular}} \vspace{-10pt}
\label{merge}
\end{table}

\subsubsection{ZoneFL User Training Time Overhead} 

\begin{figure}[t!]
\centering
    \begin{minipage}[b]{0.232\textwidth}
    \resizebox{1\textwidth}{!}{%
\includegraphics[width=0.99\linewidth,scale=1]{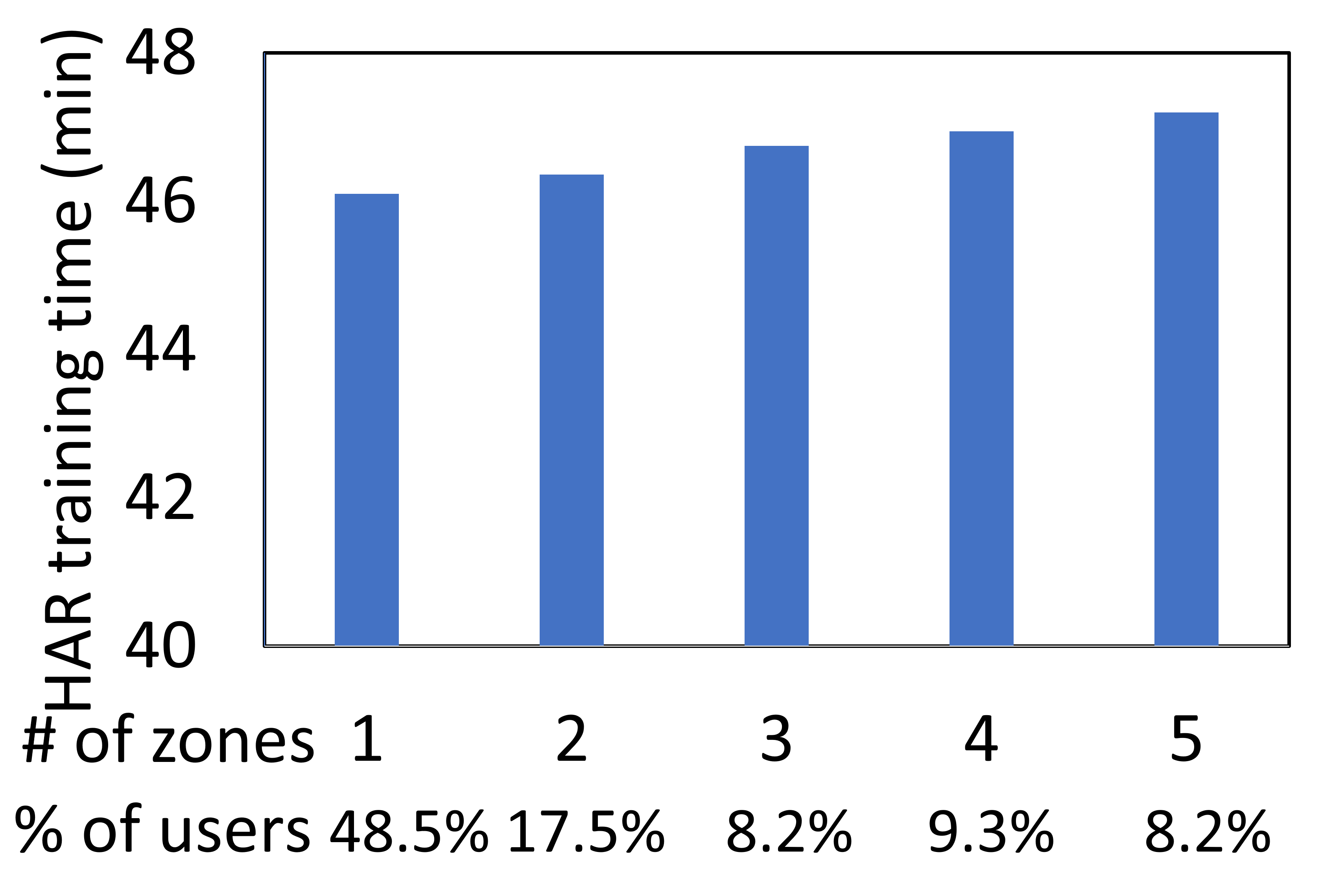}}

  \end{minipage}\vspace{-5pt}
   \hfill
  \begin{minipage}[b]{0.232\textwidth}
  \resizebox{1\textwidth}{!}{
\includegraphics[width=0.99\linewidth,scale=1]{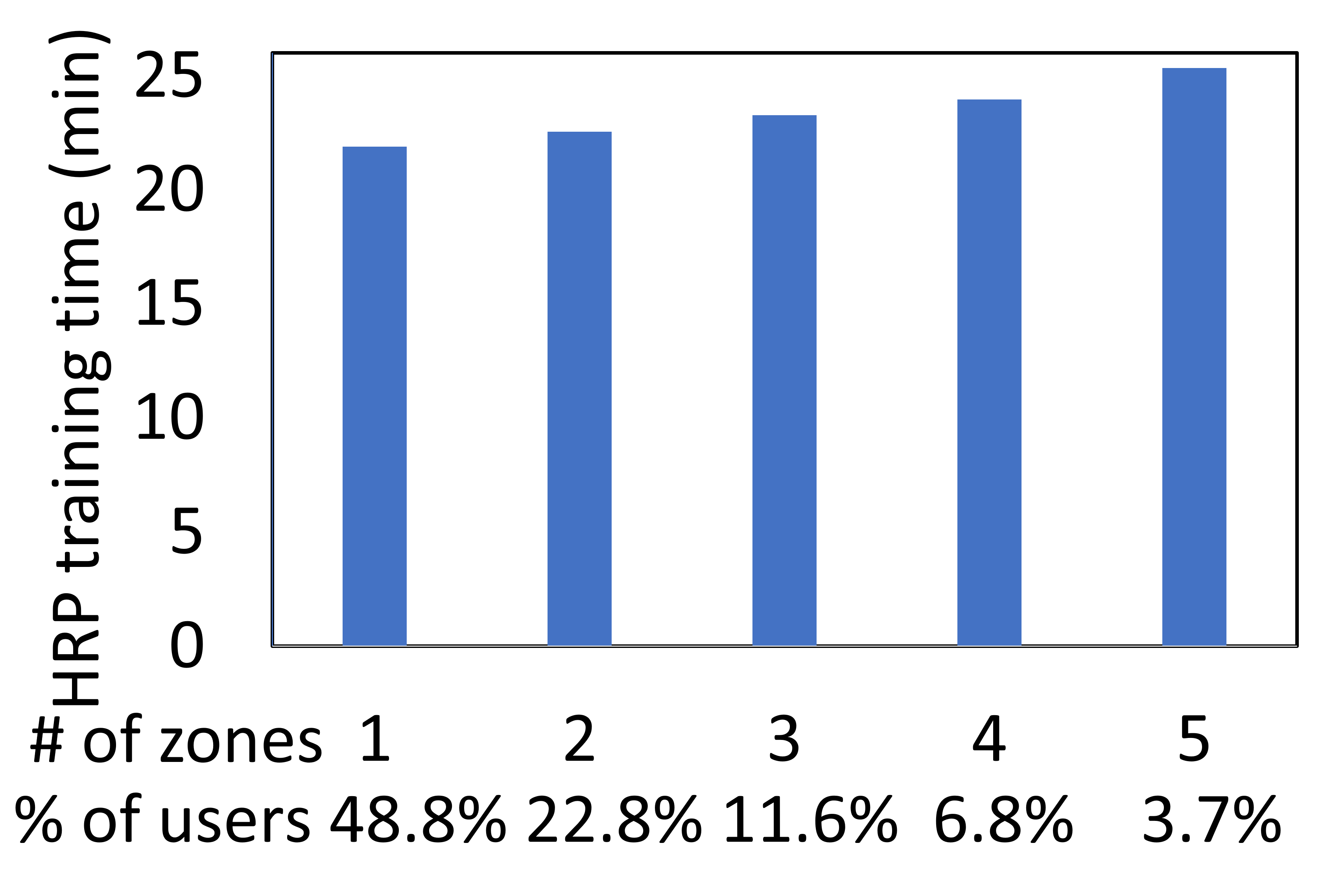}}
  \end{minipage}\vspace{-5pt}
\caption{User Training Time vs. Number of Zones in the User Data.} \vspace{5pt}
\label{fig:overhead}
\end{figure}

In ZoneFL, the phones may have data from and may train in multiple zones, which may introduce a certain level of overhead compared with Global FL. For every round in Global FL, a phone trains once for all its data. In ZoneFL, a phone may train multiple times (once per zone from where it has data), but for a smaller fraction of data. Figure~\ref{fig:overhead} illustrates the background training time in Android when the phone trains the same amount of data, while varying the number of zones the data are distributed to. The percentage of users shown under the X axis represents the fraction of users that have data in [1, 5] zones (e.g., 8.2\% of users have data in 5 zones). The number of samples trained per zone follows the average reported in section~\ref{feasibility}.  For the 49\% of the users that have data in a single zone, there is no overhead compared to Global FL (i.e., train once with all the data). For the rest of the users, we observe a small overhead, which increases with the number of zones. However, the training time overhead never exceeds 3.5 minutes. Considering that the training occurs in the background, this is an acceptable overhead for the benefits of ZoneFL in terms of model utility and server scalability.

\vspace{-2.5pt}
\section{Conclusions and Future Work}
\label{sec:con}
\vspace{-2.5pt}

This paper proposed ZoneFL, a mobile-edge-cloud FL system, that distributes training across geographical zones to improve model utility and scalability compared with traditional FL. We augmented ZoneFL with two training algorithms, ZMS and ZGD, enabling zone models to adapt to changes in user mobility behavior. ZMS and ZGD can work complementary during FL training rounds, with ZMS improving model utility in the initial rounds and ZGD further improving the utility after that. Using two different models, including human activity recognition and heart rate prediction, and mobile sensing datasets collected in the wild, we showed that ZoneFL outperforms traditional FL in terms of model utility and server scalability. We implemented an Android/AWS prototype of ZoneFL with ZMS and demonstrated the feasibility of ZoneFL in real-life conditions. As future work, we will investigate how ZGD and ZMS work together to further improve model performance, and whether similarity of data distribution among zones should be considered when defining the zone neighborhood relationship.

\balance
\bibliographystyle{plain}
\bibliography{references}

\end{document}